\documentclass[10pt,twocolumn,letterpaper]{article}

\usepackage[pagenumbers]{cvpr} 

\usepackage{graphicx}
\usepackage{amsmath}
\usepackage{amssymb}
\usepackage{booktabs}
\usepackage{amsfonts}
\usepackage{bbding} 
\usepackage{color}
\usepackage{multirow}
\usepackage[accsupp]{axessibility}
\usepackage{makecell}
\usepackage{colortbl} 
\usepackage{caption}
\usepackage[table]{xcolor}

\definecolor{Gray}{gray}{0.92} 

\usepackage[pagebackref,breaklinks,colorlinks]{hyperref}

\usepackage[capitalize]{cleveref}
\crefname{section}{Sec.}{Secs.}
\Crefname{section}{Section}{Sections}
\Crefname{table}{Table}{Tables}
\crefname{table}{Tab.}{Tabs.}

\def\cvprPaperID{0001} 
\def\confName{CVPR}
\def\confYear{2024}

\begin{document}

\title{\textbf{DriVerse: Navigation World Model for Driving Simulation via Multimodal Trajectory Prompting and Motion Alignment}}


\author{
Xiaofan Li \quad Chenming Wu \quad Zhao Yang \quad Zhihao Xu \quad \\
Dingkang Liang \quad Yumeng Zhang \quad Ji Wan \quad Jun Wang \quad \\
\vspace{4.pt}
Baidu Inc. \\
\vspace{0.5em}
\url{https://github.com/shalfun/DriVerse}
}

\twocolumn[
\begin{center}  
\maketitle
\centering  
\resizebox{1\linewidth}{!}{  
\includegraphics{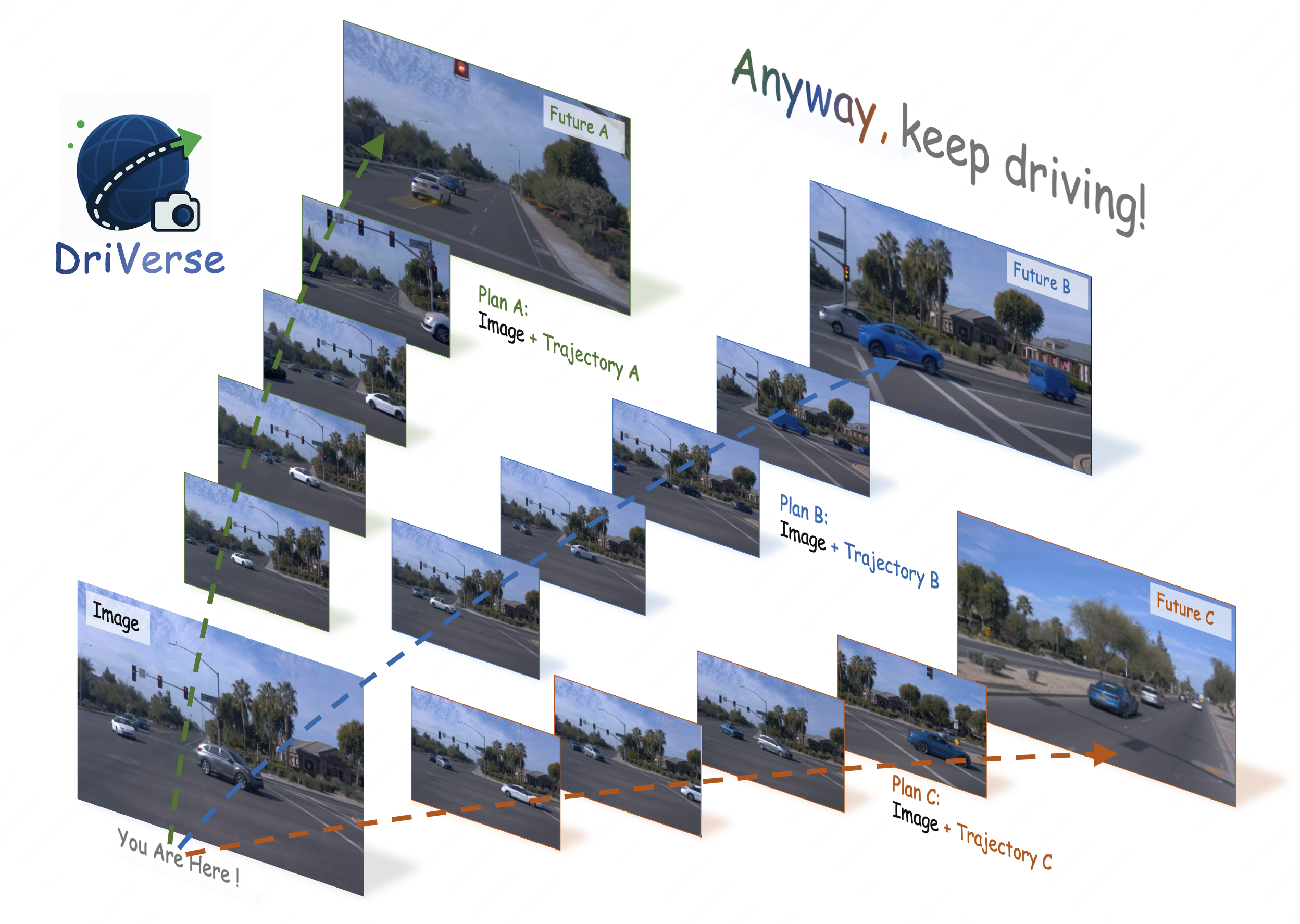}}  
\captionof{figure}{Our proposed navigation world model for driving simulation, referred to as \textit{DriVerse}, is designed to transform a single input image along with various navigation trajectories into high-quality videos that accurately reflect the intended motion, ensuring that the generated videos maintain a strong alignment with real-world driving scenarios, which significantly enhances the realism and utility of driving simulations.}  
\label{fig:m0}
\end{center}]

\begin{abstract}
This paper presents DriVerse, a generative model for simulating navigation-driven driving scenes from a single image and a future trajectory. Previous autonomous driving world models either directly feed the trajectory or discrete control signals into the generation pipeline, leading to poor alignment between the control inputs and the implicit features of the 2D base generative model, which results in low-fidelity video outputs. Some methods use coarse textual commands or discrete vehicle control signals, which lack the precision to guide fine-grained, trajectory-specific video generation, making them unsuitable for evaluating actual autonomous driving algorithms. DriVerse introduces explicit trajectory guidance in two complementary forms: it tokenizes trajectories into textual prompts using a predefined trend vocabulary for seamless language integration, and converts 3D trajectories into 2D spatial motion priors to enhance control over static content within the driving scene. To better handle dynamic objects, we further introduce a lightweight motion alignment module, which focuses on the inter-frame consistency of dynamic pixels, significantly enhancing the temporal coherence of moving elements over long sequences.
We also propose an inference-time strategy to address issues caused by rapid vehicle heading changes.
With minimal training and no need for additional data, DriVerse outperforms specialized models on future video generation tasks across both the nuScenes and Waymo datasets. The code and models will be released to the public.
\end{abstract}

\section{Introduction}

Driving World Models (DWMs) are essential for autonomous driving, allowing vehicles to forecast future scenes based on present or historical observations. These models enable self-driving systems to foresee potential hazards, simulate diverse scenarios, and make informed decisions. In addition, DWMs can augment training data, especially for rare or safety-critical situations, and provide realistic simulated environments to support end-to-end reinforcement learning. 

Recent advances in image and video generation have propelled the development of generative models for autonomous driving. Early efforts often adapted open-source models pre-trained on large-scale 2D datasets, transferring their powerful capabilities to the driving domain. A line of works leverage the layout for generation, such as DriveDreamer~\cite{wang2023drivedreamer}, DrivingDiffusion~\cite{li2024drivingdiffusion}, MagicDrive~\cite{gao2023magicdrive}, Drive-WM~\cite{wang2024driving}, WoVoGen~\cite{lu2024wovogen}, HoloDrive~\cite{wu2024holodrive}, and DrivePhysica~\cite{yang2024physical}, utilize the layout as a conditional input to guide video generation. However, it is impractical to assume full access to layout annotations in all scenes when simulating real-world scenarios. The strong dependence on detailed layout inputs limits their application in closed-loop simulation.

Other approaches like GAIA-1~\cite{hu2023gaia}, MUVO~\cite{bogdoll2023muvo}, ADriver-I~\cite{jia2023adriver}, DriveDreamer-2~\cite{zhao2024drivedreamer4d}, DriveWorld~\cite{min2024driveworld}, BEVWorld~\cite{zhang2024bevworld}, MPDrive~\cite{zhang2025mpdrive}, WorldGPT~\cite{ge2024worldgpt} and HERMES~\cite{zhou2025hermes} rely on text prompts or discrete control signals. While effective for open-ended generation, they lack precise control over the generation process for specific trajectories, making them less suitable for evaluation or planning under strict constraints. Furthermore, trajectory-conditioned models such as Vista~\cite{gao2024vista}, OccWorld~\cite{zheng2024occworld}, OccSora~\cite{wang2024occsora}, DOME~\cite{gu2024dome}, and Imagine-2-Drive~\cite{garg2024imagine} attempt to encode 3D trajectories as generation conditions. However, directly encoding 3D coordinates or projecting them onto 2D planes often results in poor alignment with pre-trained generative features. This misalignment makes it difficult to learn from limited data and can degrade the generalization capability of the underlying generative backbone. As a result, these methods struggle to produce temporally consistent videos in challenging scenarios such as turning or lane changing. 

To tackle the challenges mentioned, we introduce DriVerse, a trajectory-guided video generative world simulator specifically designed for navigation-focused autonomous driving scenarios. DriVerse takes a single scene image and its future trajectory to produce high-quality driving videos that align closely with the specified path.
At the heart of DriVerse is a multimodal trajectory prompting (MTP) strategy that utilizes both semantic and spatial cues to steer video generation. We define a trend vocabulary that converts the trajectory's motion patterns into a sequence of discrete tokens, which are added to traditional text prompts for enhanced semantic conditioning. Additionally, we project the 3D trajectory into 2D image space by translating it into a series of anchor point motions. This approach aligns navigation control signals with image-based movements, effectively transforming abstract vehicle trajectories into pixel-level motion paths. The control signal represents the motion trend of static 3D anchors as they are projected onto image pixels over time as the vehicle progresses.


In particular, we observe that current 2D image-to-video generation models often struggle with complex outdoor scenes, notably those featuring high-frequency visual details and dense dynamic elements (e.g., vehicles, pedestrians). This can lead to physically implausible behaviors. To address this, we introduce LMA (Latent Motion Alignment), a lightweight module that explicitly models inter-frame dynamics in the latent space. LMA tracks the motion of dynamic regions across frames and propagates gradient feedback during training, enhancing temporal consistency and the physical realism of dynamic objects over extended sequences.

Furthermore, most of exisiting video generation backbone models are trained under the assumption of gradual scene transitions and minimal camera motion. However, in autonomous driving scenarios—such as sharp turns, U-turns, or complex intersections—the vehicle's heading can change dramatically. This can render pixel information from the initial frame quickly obsolete, leading to significant temporal artifacts and scene inconsistencies in subsequent frames. To overcome this challenge, we propose a Dynamic Window Generation (DWG) strategy, which adaptively refreshes the conditioning frame during autoregressive extension at inference time. DWG monitors changes in the heading angle and updates the generation window at appropriate intermediate frames.
DriVerse substantially outperforms previous models in both visual quality and trajectory fidelity, especially when generating long-horizon future videos conditioned on diverse trajectories, as illustrated in Fig.~\ref{fig:m0}.
Our main contributions are summarized as follows: 
1) We propose DriVerse, a trajectory-guided driving world simulation model that achieves both high-quality generation and strong generalization without relying on large-scale training data.
2) We design MTP (Multimodal Trajectory Prompting), which encodes the trajectory as both a sequence of discrete trend tokens—derived from a predefined motion vocabulary—and the motion of 3D static anchors in 2D space. In addition, we introduce LMA (Latent Motion Alignment) to enhance control precision over static scenes and ensure temporal consistency for dynamic objects.
3) DriVerse demonstrates strong performance on the nuScenes and Waymo datasets, achieving the best results compared to existing driving world models.


\section{Related Work}
\subsection{Video Generation Models}

With advances in training stability and strong backing from open-source communities, recent developments in video generation have been significantly propelled by diffusion models~\cite{ho2020denoising, song2020denoising, peebles2023scalable}. Many contemporary works adopt the text-to-video (T2V) paradigm~\cite{karras2023dreampose, ruan2023mm, zhang2023i2vgen, latentvideodiffusion, chen2023videocrafter1, hong2022cogvideo, yang2024cogvideox}, while others shift their focus from text to images, emphasizing the image-to-video (I2V) approach~\cite{mcdiff, chen2023seine, esser2023structure}. The Video Diffusion Model~\cite{ho2022video} was among the pioneers to adapt a 2D diffusion architecture for video, training it jointly on both image and video data from the ground up.

To effectively utilize powerful pre-trained image generators like Stable Diffusion~\cite{ldm}, subsequent works have integrated temporal modules into the 2D backbone and fine-tuned them on large-scale video datasets~\cite{webvid10m}. Other notable directions in this field include scalable Transformer-based architectures~\cite{ma2024latte}, modeling in spatiotemporally compressed latent spaces (as seen in W.A.L.T.\cite{gupta2023photorealistic} and Sora\cite{videoworldsimulators2024}), and discrete token generation guided by language models~\cite{kondratyuk2023videopoet, chen2024egocentric}. Effective factors in visual in-context learning have also been analyzed in detail~\cite{sun2025exploring}. For a more detailed overview, we recommend readers consult~\cite{Po:2023:star_diffusion_models}. Recently, models like Stable Video Diffusion (SVD)\cite{svd}, based on U-Net, and Wan2.1\cite{wan2025}, leveraging Diffusion Transformer (DiT) architecture for better scale-up performance, have further refined the pipeline with enhanced training methodologies and data selection strategies.

In the realm of controllable video generation, several studies have integrated additional control signals, such as depth maps and pose sequences, to precisely guide scene and human movements~\cite{sparsectrl, chen2023control, zhang2023controlvideo, khachatryan2023text2video, hu2023animate, xu2023magicanimate}. Complementary to diffusion video generators, autoregressive next-focus prediction over images~\cite{li2025fvar} and the framing of image editing as a temporally degenerate process~\cite{li2025video4edit} define alternative inductive biases for structured appearance change, while 3D Gaussian score-distillation editing~\cite{qu2025drag} and NeRF-based multi-view object detection indoors~\cite{huang2025nerf} stress cross-view geometry in settings complementary to monocular I2V on streets. Visual reference prompts for foundation segmenters (e.g., {SAM}~\cite{sun2024vrpsam}) further highlight how non-textual conditions steer large vision models. For example, SparseCtrl~\cite{sparsectrl} employs sketch images as spatial control inputs, resulting in enhanced video quality and improved temporal coherence. In terms of camera control, AnimateDiff utilizes lightweight LoRA-based fine-tuning~\cite{hu2021lora} to accommodate various types of camera movements.

Direct-a-Video~\cite{directavideo} introduces a camera embedding module to manage camera poses, but it is limited to only three basic parameters (e.g., pan-left), which constrains its control capabilities. Meanwhile, MotionCtrl~\cite{motionctrl} and CameraCtrl~\cite{he2024cameractrl} broaden the range of parameters available for viewpoint control; however, they still face challenges in providing fine-grained pixel-level motion supervision. This level of precision is particularly critical in fields like autonomous driving, where spatial accuracy and trajectory alignment are essential for safety and effectiveness.

Furthermore, while these methods demonstrate some level of generalization, their pre-training datasets primarily comprise indoor environments or relatively simple outdoor scenes characterized by limited camera motion and few dynamic elements. When applied directly to autonomous driving scenarios, such models often struggle to accurately capture the implicit scene graph of real-world street environments, have difficulty modeling the movements of vehicles and pedestrians, and fail to reproduce high-frequency details effectively. In contrast, DriVerse explicitly prioritizes the modeling of dynamic objects in urban street scenes, effectively addressing these limitations and enhancing the applicability of diffusion-based video generation for driving applications.

\subsection{Navigation World Models for Driving}

World models aim to predict future scene evolution based on current observations~\cite{ha2018world, guan2024world, tu2025role}, and have increasingly become a research focus in autonomous driving. Many approaches~\cite{kim2021drivegan, lu2024wovogen, li2024drivingdiffusion, wen2024panacea, yang2025dualdiff} focus on predicting 2D visual representations, demonstrating strong capabilities in dynamic modeling. Closely related, learning multiple probabilistic decisions from a latent world model~\cite{xiao2025learning} further links long-horizon rollouts to action diversity. In parallel, vision--language research revisits MLLM image classification under classic benchmarks~\cite{liu2024revisitingmllms}, improves descriptive captioning for multimodal perception with specialist models~\cite{sun2024descriptivecaption}, explores grow-and-refine memory for agentic multimodal learning~\cite{bo2025agenticlearner}, and mitigates visual forgetting in efficient MLLMs via resampling~\cite{feng2025visionremember}---all adjacent trends for scalable conditioning interfaces in embodied and automotive AI, though with objectives beyond trajectory-consistent video.

Most mainstream methods are based on generative models, primarily employing autoregressive Transformers~\cite{hu2023gaia, chen2024drivinggpt} or diffusion models~\cite{guo2024infinitydrive, jiang2024dive, wen2024panacea_plus} for future video prediction. Specifically, GAIA-1~\cite{hu2023gaia} formulates world modeling as a sequence generation task using an autoregressive Transformer; ADriver-I~\cite{jia2023adriveri} combines multi-modal large language models (LLMs) and diffusion for joint control and frame generation. DriveDreamer~\cite{wang2024drivedreamer} incorporates structured traffic constraints into the diffusion process, while DriveDreamer-2~\cite{zhao2024drivedreamer4d} further leverages LLMs for personalized video generation. Drive-WM~\cite{wang2024drive_wm} enhances multi-view consistency via view factorization, and GenAD~\cite{yang2024genad} improves zero-shot generalization through large-scale video datasets.

However, these world models often rely on high-level semantic reasoning and struggle to produce physically plausible, detail-rich long-horizon videos when ego-vehicle behaviors change drastically (e.g., turning or lane changing). Vista~\cite{gao2024vista}, built upon GenAD, incorporates extensive additional data and introduces structural and motion losses to improve dynamic modeling and structural fidelity, achieving high-resolution, high-quality long-term scene generation. Subsequent works further advance the duration of predictions~\cite{guo2024infinitydrive, hu2024drivingworld} and unify prediction with planning~\cite{chen2024drivinggpt, wang2024driving} or perception and localization~\cite{liang2025seeing, li2025u}.

Despite Vista's impressive performance—enabled by large-scale training—its strong bias toward straight-driving samples in the training set leads to overfitting. This results in degradation in video quality under more complex scenarios such as left/right turns. In contrast, DriVerse achieves both high generation quality and strong generalization using only limited training data, demonstrating greater robustness and adaptability in diverse driving scenes.


\begin{figure*}[t]
	\begin{center}
		\includegraphics[width=1.\linewidth]{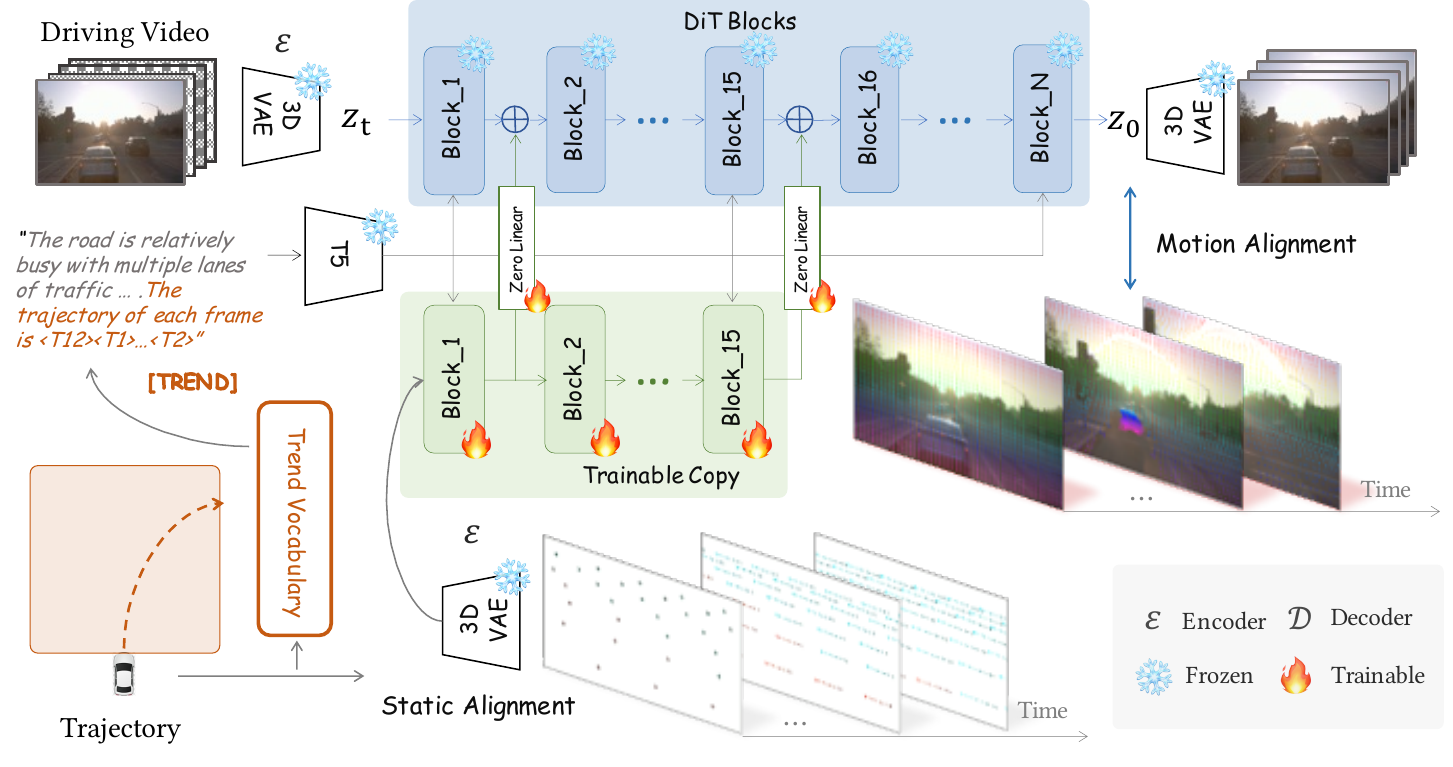}
	\end{center}
    \caption{Overview of the DriVerse framework. Given a single scene image and a future trajectory, DriVerse decomposes the generation task into static alignment and motion alignment. The former, referred to as Multimodal Trajectory Prompting (MTP), encodes future trajectories into textual prompts using a predefined trend vocabulary, and further injects spatial motion priors derived from 3D static anchors into the frozen backbone via a trainable control module. The latter, called Latent Motion Alignment (LMA), supervises the generation by enforcing consistency between generated and ground-truth dynamic pixels, based on offline-computed motion correspondences.}
	\label{fig:m1}
\end{figure*}

\section{Proposed Method}

\noindent{\textbf{Problem Formulation. }} 
Given an input image $I_0$ and a future navigation trajectory $\tau = \{p_1, p_2, ..., p_T\}$ defined in the 3D ego-vehicle coordinate space, our objective is to synthesize a video $\mathcal{V} = \{I_1, I_2, ..., I_T\}$ such that each frame $I_t$ aligns with the intended vehicle motion and dynamic scene evolution implied by the trajectory. We assume no access to any future frame at the phase of inference and do not rely on LiDAR, layout, or additional control signals.

\noindent{\textbf{Overview. }} 
We adopt the block design of DiT in our backbone for latent-based video generation, following the scalable transformer-based diffusion architecture proposed in~\cite{peebles2022dit}. To improve controllability over generation, we draw inspiration from PixArt-$\alpha$~\cite{zhang2023pixart}, which integrates trajectory-relevant prompt information into DiT via cross-attention and trend-aware tokenization.
Building upon this foundation, our proposed framework—DriVerse—introduces two key mechanisms for trajectory-conditioned generation:
(1) \textit{Multimodal Trajectory Prompting (MTP):} a dual-mode prompting scheme that encodes future trajectories as both textual prompts and 3D static anchors to control static scene content; and
(2) \textit{Latent Motion Alignment (LMA):} a lightweight alignment module that supervises inter-frame consistency of dynamic objects by enforcing motion-aware latent consistency.
An overview of our framework is illustrated in Fig.~\ref{fig:m1}.

\subsection{\textbf{Multimodal Trajectory Prompting} \label{m31}}
To enable precise control over the generated video, we introduce a multimodal trajectory prompting mechanism that encodes the 3D trajectory in two complementary forms: 1) a high-level symbolic representation, using a trend vocabulary in the language modality, and 2) a low-level spatial prior, represented by trajectory-guided spatial anchors in the 3D point modality.

\noindent{\textbf{Trend Tokenization of 3D Trajectories.}}
To encode directional priors of future trajectories, we discretize each 3D trajectory segment into a predefined trend vocabulary. Specifically, the $360^\circ$ horizontal plane is uniformly divided into 12 angular sectors, each corresponding to a directional token $\texttt{<T}_k\texttt{>}$, where $k \in \{1, \dots, 12\}$ denotes a clock-hour direction.
Given a trajectory $\{\mathbf{x}_t\}_{t=1}^T$, we compute segment directions $\Delta \mathbf{x}_t = \mathbf{x}_{t+1} - \mathbf{x}_t$ and map each to its token via the quantized azimuthal angle $\theta_t$:
\[
\texttt{token}_t = \texttt{<T}_{\mathrm{quantize}(\theta_t)}\texttt{>} \quad
\theta_t = \angle(\Delta \mathbf{x}_t).
\]
To ensure compatibility with frozen language models and avoid introducing new special tokens, we embed all trend tokens into the prompt using natural language definitions. Specifically, we define their semantics inline at the beginning of the prompt using a fixed textual template:
\begin{quote}
\texttt{"<T1> to <T12> represent the 12 clock directions, each indicating a different heading angle. I will use them to describe the trajectory: the trajectory of each frame is \{\}."}
\end{quote}
This approach allows explicit control injection while preserving the integrity of pretrained language priors.

\noindent{\textbf{Trajectory-Guided Spatial Anchors. }}
To provide spatial guidance for modeling static content in the driving scene, we introduce Trajectory-Guided Spatial Anchors (TSA). For each video sequence, we first construct a set of static 3D anchor points $\{\mathbf{X}_j^0\}_{j=1}^K$ in the world coordinate system, centered around the ego vehicle's position at the first frame. These anchors are uniformly distributed within a fixed radius and remain static across time.

Each anchor $\mathbf{X}_j^0$ is projected to the image plane of the first frame using the camera intrinsics $\mathbf{K}$ and the ego-pose transformation matrix $\mathbf{T}_0$ as follows:
\begin{equation}
    \mathbf{u}_j^0 = \pi\left(\mathbf{K} \cdot \mathbf{T}_0 \cdot \mathbf{X}_j^0\right),
\end{equation}
where $\pi(\cdot)$ denotes the standard perspective projection, and $\mathbf{u}_j^0$ is the resulting 2D pixel coordinate.

For each subsequent frame $t > 0$, the 3D position of each anchor is updated relative to the current ego-pose $\mathbf{T}_t$, resulting in:
\begin{equation}
    \mathbf{u}_j^t = \pi\left(\mathbf{K} \cdot \mathbf{T}_t \cdot \mathbf{X}_j^0\right).
\end{equation}

To enrich temporal context, we additionally project anchors from the past $M$ frames into the current frame, yielding a set of trailing projections $\{\mathbf{u}_j^{t-m}\}_{m=1}^M$. Each projected point is assigned a transparency value based on its displacement from the current projection:
\begin{equation}
    \delta_j^{t-m} = \left\| \mathbf{u}_j^t - \mathbf{u}_j^{t-m} \right\|_2 \quad
    \alpha_j^{t-m} = \exp(-\lambda \cdot \delta_j^{t-m}),
\end{equation}
where $\lambda$ is a temperature parameter controlling the decay rate. This process creates a temporal fading trail for each anchor that encodes its projected motion history.

Furthermore, we compute the 2D motion vector of each anchor between consecutive frames as:
\begin{equation}
    \mathbf{v}_j^t = \mathbf{u}_j^t - \mathbf{u}_j^{t-1},
\end{equation}
and apply a color mapping function $\mathcal{C}(\cdot)$ to visualize the direction and magnitude of motion in a perceptually meaningful manner, similar to optical flow color encodings:
\begin{equation}
    \mathbf{c}_j^t = \mathcal{C}(\mathbf{v}_j^t).
\end{equation}

These colored anchor trajectories serve as interpretable motion-aware priors that enhance the model's ability to obey structured spatial-temporal consistency.

\begin{figure}[t]
\centering
    \includegraphics[width=\columnwidth]{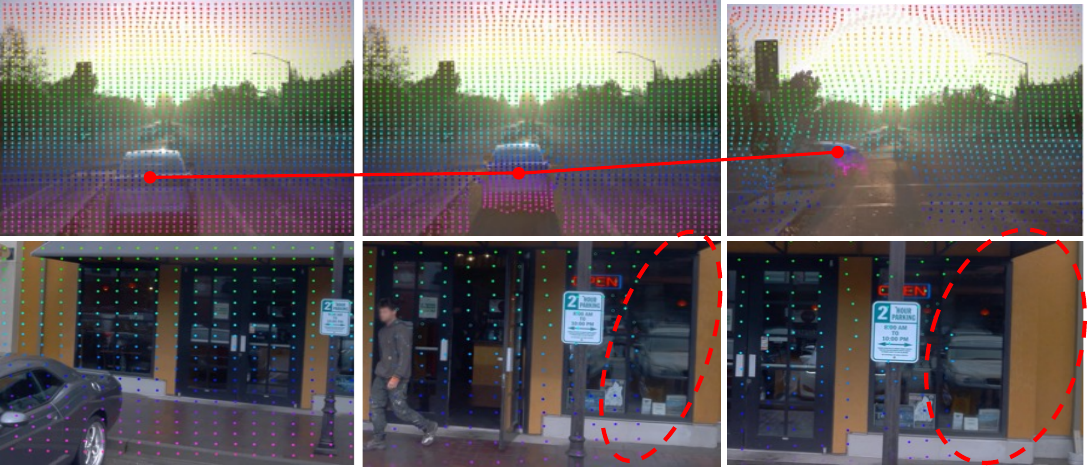}
  \caption{The top and bottom parts of the figure visualize, through pixel-level tracking, the respective influences of object motion and camera motion on image pixels. The red-circled regions highlight cases where a large change in the ego vehicle's heading angle leads to a reduction in the number of initialized static points. }
  \label{fig:m3}
\end{figure}


\subsection{Latent Motion Alignment \label{m32}}
To provide structurally consistent motion supervision for video generation, we employ CoTracker v3~\cite{karaev24cotracker3} to extract pixel-wise trajectories across frames. CoTracker leverages a transformer-based architecture to predict long-range point correspondences without requiring explicit optical flow or object detection. To focus on dynamic regions, we compute motion magnitude masks from the tracked trajectories and uniformly sample a set of spatial points $\{p_i^0\}_{i=1}^N$ within regions of large motion in the first frame.

For each sampled point $p_i^0$ in the first frame, we obtain its corresponding locations $\{p_i^t\}_{t=1}^T$ in subsequent frames. Rather than computing RGB consistency in pixel space, we measure latent consistency directly in the VAE's intermediate feature space before decoding. Let $z_t$ denote the latent feature map at time $t$, and $z_t(p_i^t)$ the feature vector at the tracked location. We define a motion-weighted consistency loss as:
\[
\mathcal{L}_{\text{cons}} = \frac{1}{N} \sum_{i=1}^{N} w_i \sum_{t=1}^{T} \left\| z_t(p_i^t) - z_0(p_i^0) \right\|_2^2,
\]

where $w_i$ is a motion-dependent weight computed as the normalized displacement magnitude:
\[
w_i = \frac{\sum_{t=1}^{T} \left\| p_i^t - p_i^0 \right\|_2}{\sum_{j=1}^{N} \sum_{t=1}^{T} \left\| p_j^t - p_j^0 \right\|_2}.
\]

This design ensures that trajectories with larger motion contribute more significantly to the consistency loss, encouraging the model to maintain stable latent representations in highly dynamic regions.

\subsection{Dynamic Window-based Generation \label{m33}}
Following the inference approach proposed in~\cite{li2024drivingdiffusion}, our video generation employs a combination of diffusion-based inference and autoregressive extension. Formally, given an initial frame \( I_0 \) and trajectory condition \( T \), the model generates an initial sequence of \( N \) frames, followed by autoregressive generation using a sliding window:
\[ \{I_1, I_2, \dots, I_{k+N}\} = G(I_k, T_{k:k+N}) \quad \text{for } k = 0, N, 2N, \dots \]
where \( G(\cdot) \) denotes the generative process of the model.
However, in scenarios with rapid changes in the vehicle's heading, pixel-level information from the initial frame quickly becomes obsolete, as shown in Figure~\ref{fig:m3},. To address this, we introduce the Dynamic Window Generation (DWG) strategy. Specifically, rather than always using the last generated frame \( I_N \) for autoregressive expansion, DWG adaptively selects a suitable intermediate frame based on the rate of heading-angle change and the visibility of predefined 3D anchor points.

Specifically, we leverage the anchor points introduced in \textit{Trajectory-Guided Spatial Anchors} to assess scene stability. Let $\mathcal{A}_0$ denote the set of anchor points in the initial frame and $\mathcal{A}_t$ the subset that remains visible in frame $I_t$. We define anchor visibility as:
\[ V_t = \frac{|\mathcal{A}_t|}{|\mathcal{A}_0|} \]
where \( |\mathcal{A}_t| \) is the number of anchors remaining visible in frame \( I_t \). The DWG frame selection criterion is then defined as follows:
\[I_{\text{key}} = \begin{cases}
I_N, & \text{if } V_t \geq 0.6 \quad \forall t \in \{1,\dots,N\}\\[5pt]
I_{t^*}, & \text{otherwise, where } t^* = \arg\min_t \{V_t < 0.6\}
\end{cases}\]

In practice, if at least 60\% of the anchors remain visible throughout the generated window, we continue autoregressive generation from the last frame. Otherwise, the earliest frame at which visibility falls below 60\% serves as the new conditioning frame for subsequent video generation. This adaptive mechanism effectively maintains scene consistency and reduces temporal artifacts in scenarios involving rapid directional changes.

\begin{figure*}[t]
	\begin{center}
		\includegraphics[width=1.00\linewidth]{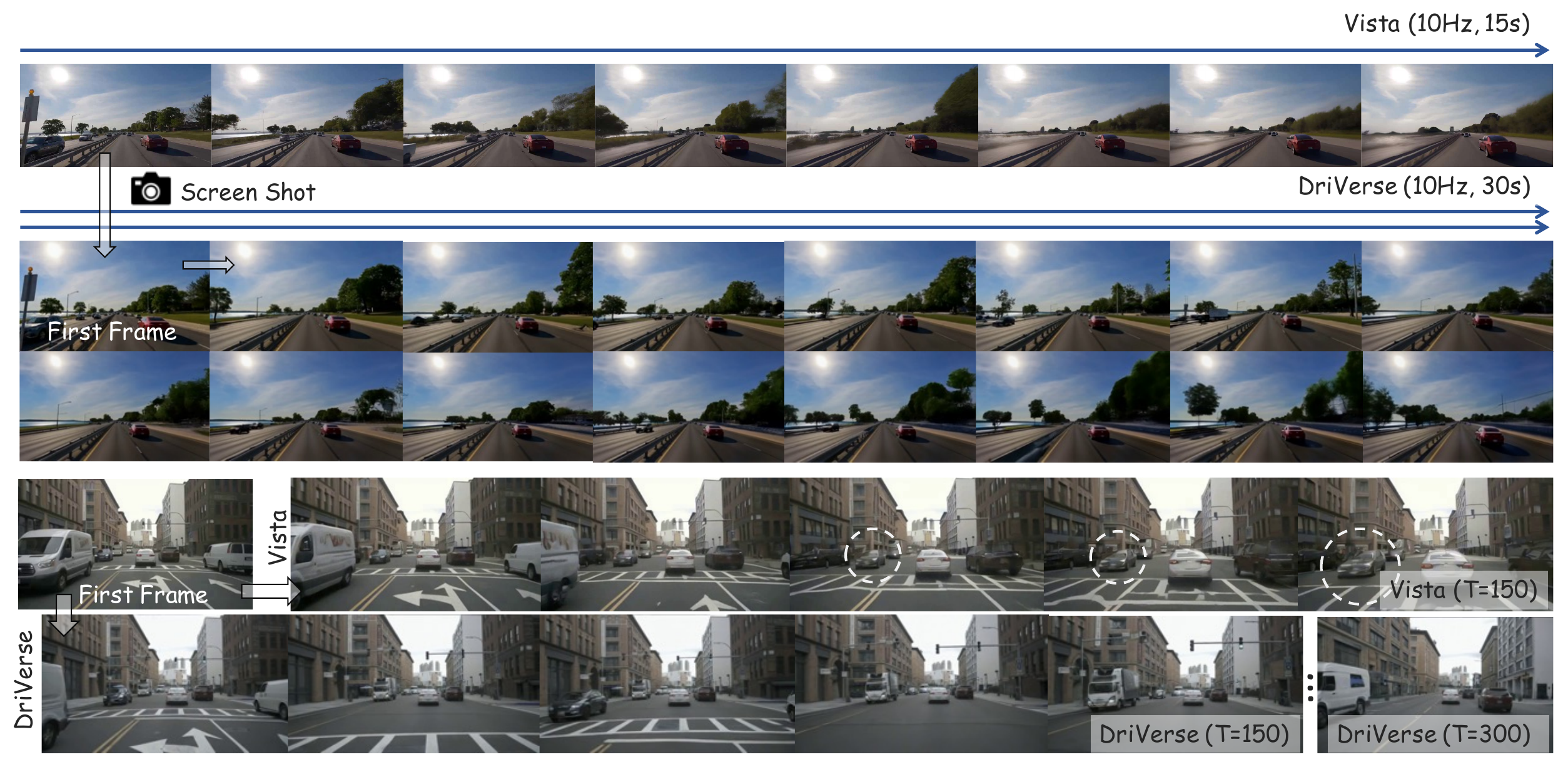}
	\end{center}
         \vspace{-8pt}
    \caption{Qualitative comparison with existing methods. \textbf{Top:} Visualization adapted from the original Vista paper. We input only the first frame to DriVerse, which is capable of generating high-quality, long-horizon future predictions. \textbf{Bottom:} Comparison between DriVerse and Vista. White dashed circles indicate regions of implausible generation.}
	\label{fig:e1}
\end{figure*}

\begin{table*}[t]
    \footnotesize
    \centering
    \caption{Comparison of our method with specialized generation models on nuScenes and Waymo Open Dataset. $^{\bigstar}$ marks our baseline method, which is fine-tuned under the same resolution and iteration constraints using the official pre-trained weight of Vista~\cite{gao2024vista}. Bold indicates the best performance. }
    \setlength{\tabcolsep}{4.8mm}
    
    \begin{tabular}{lcccc|cc}
    \toprule
    Method & Reference & Dataset & Resolution & Max Video Length & FID $\downarrow$ & FVD $\downarrow$ \\
    \midrule
    DriveGAN~\cite{kim2021drivegan} & CVPR 21 & nuScenes & 256$\times$256 & -- & 73.4 & 502.3 \\
    DriveDreamer~\cite{wang2024drivedreamer} & ECCV 24 & nuScenes & 128$\times$192 & 12Hz, 4s & 52.6 & 452.0 \\
    WoVoGen~\cite{lu2024wovogen} & ECCV 24 & nuScenes & 256$\times$448 & 2Hz, 2.5s & 27.6 & 417.7 \\
    GenAD~\cite{yang2024genad} & CVPR 24 & nuScenes & 256$\times$448 & 2Hz, 4s & 15.4 & 184.0 \\
    Drive-WM~\cite{wang2024driving} & CVPR 24 & nuScenes & 192$\times$384 & 2Hz, 8s & 15.2 & 122.7 \\
    Vista$^{\bigstar}$~\cite{gao2024vista} & NeurIPS 24 & nuScenes & 480$\times$832  & 10Hz, 15s & 23.4 & 158.0 \\
    \rowcolor{gray!10}
    DriVerse (Ours) & -- & nuScenes & 480$\times$832 & 10Hz, 8s & \textbf{18.2} & \textbf{95.2} \\
    \rowcolor{gray!10}
    DriVerse (Ours) & -- & nuScenes & 480$\times$832 & 10Hz, 15s & 20.1 & 143.5 \\
    \midrule
    Vista$^{\bigstar}$ & NeurIPS 24 & Waymo Open Dataset & 480$\times$832  & 10Hz, 15s & 29.2 & 285.7 \\
    \rowcolor{gray!10}
    DriVerse (Ours) & -- & Waymo Open Dataset & 480$\times$832 & 10Hz, 8s & \textbf{25.0} & \textbf{204.3} \\
    \rowcolor{gray!10}
    DriVerse (Ours) & -- & Waymo Open Dataset & 480$\times$832 & 10Hz, 15s & 28.4 & 264.5 \\
    \rowcolor{gray!10}
    DriVerse (Ours) & -- & Waymo Open Dataset & 480$\times$832 & 10Hz, 30s & 31.3 & 281.4 \\
    \bottomrule 
    \end{tabular}
    \label{tab:main_results}
\vspace{-8pt}
\end{table*}

\section{Experiments}

\subsection{Dataset and Evaluation Metric}
\noindent{\textbf{Datasets.}}
We conduct experiments on two large-scale autonomous driving benchmarks: nuScenes~\cite{caesar2020nuscenes} and the Waymo Open Dataset (WOD)~\cite{sun2020scalability}. Both datasets provide multi-modal sensor data, including RGB images, LiDAR point clouds, and scene annotations. The nuScenes dataset contains 1,000 driving sequences collected in diverse urban environments, each lasting 20 seconds, with camera frames recorded at 12 Hz. WOD consists of 2,030 driving segments, each also 20 seconds long, captured at a frame rate of 10 Hz, and covers a broader range of road types and traffic scenarios. Following Vista~\cite{gao2024vista}, we use the front-camera RGB images as our primary training data.

\noindent{\textbf{Evaluation metric.}} Vista is the only publicly available DWM, and fortunately, it demonstrates state-of-the-art performance, we use its official results on nuScenes as the primary quantitative baseline. For qualitative comparisons, we also align our evaluation settings with Vista and conduct side-by-side visual comparisons.
We evaluate on 5,369 frames from the nuScenes validation set and 2,000 frames from the Waymo Open Dataset (WOD). The 2,000 frames from WOD are selected to ensure a uniform distribution over driving trajectories.
To assess the visual quality and temporal coherence of generated videos, we report Fréchet Inception Distance (FID)~\cite{heusel2017gans} and Fréchet Video Distance (FVD)~\cite{unterthiner2019fvd}. Lower scores on both metrics indicate better generation quality.
Beyond standard perceptual metrics, we further introduce a geometric consistency evaluation to specifically assess the model's ability to follow the given trajectory in 3D space. This metric reflects how well the generated video preserves the intended motion path. Concretely, we reconstruct local segments of the generated video and estimate their relative camera poses. These predicted poses are then aligned to the ground-truth trajectory using a similarity transformation, and the resulting alignment error quantifies the trajectory-following accuracy. Details are described in Section~\ref{subsec:e1}.

\begin{figure*}[t]
	\begin{center}
		\includegraphics[width=1.0\linewidth]{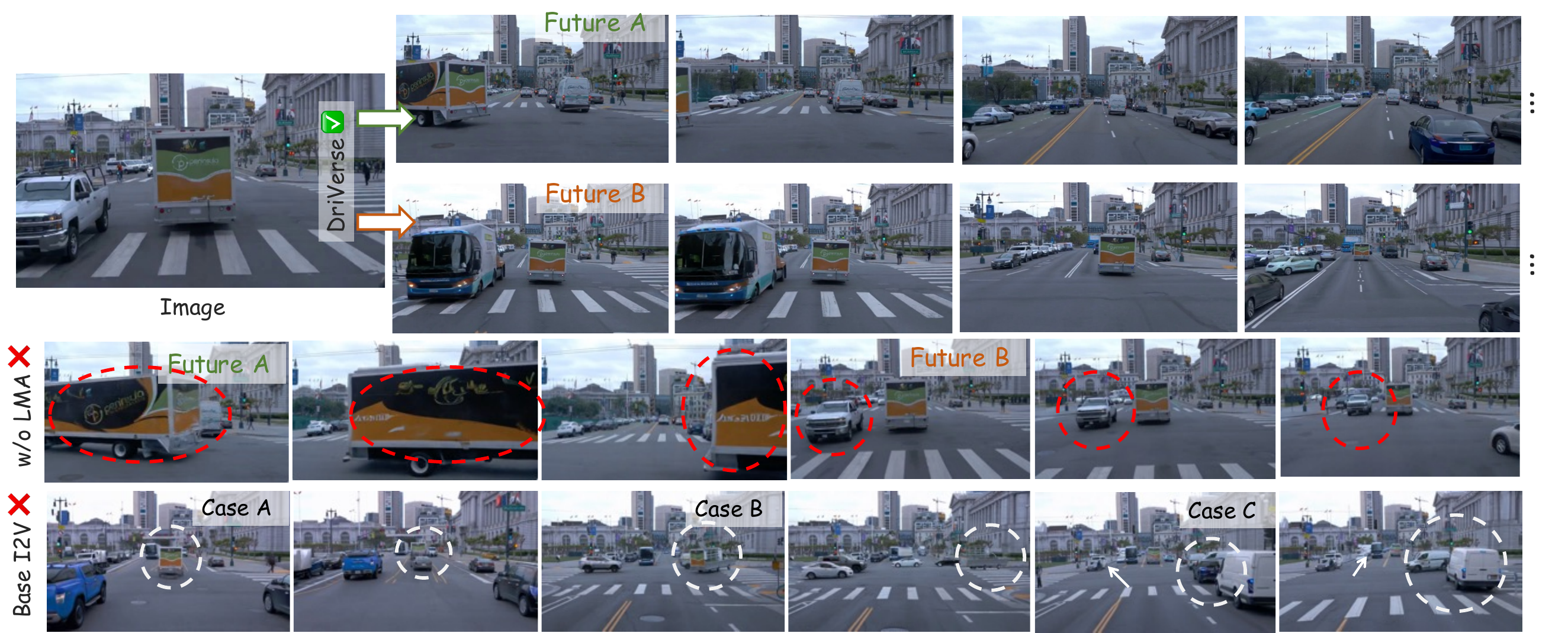}
	\end{center}
         \vspace{-4pt}
    \caption{Qualitative comparison of inference results on the Waymo Open Dataset. The top two rows show the diverse future predictions (Future A/B) generated by the full DriVerse model. The third row presents the results of the model without the LMA module, where red dashed circles highlight regions with inconsistent or implausible motion. The fourth row shows the generation results of a base image-to-video (I2V) model applied directly to street scenes, where white dashed circles and arrows indicate noticeable visual artifacts. }
	\label{fig:e2}
\end{figure*}

\subsection{Main Results}
\label{subsec:e1}

Due to page limitations, we provide the training details in the supplementary material and present the main results in this section.

\noindent{\textbf{Comparison with Specialized Methods.}}
Table~\ref{tab:main_results} presents a quantitative comparison between our method and existing state-of-the-art video generation models on both the nuScenes and Waymo Open Dataset benchmarks. On nuScenes, DriVerse consistently achieves the best performance across all settings, significantly outperforming previous methods in both FID and FVD. 
DriVerse maintains a performance advantage over the strong Vista baseline~\cite{gao2024vista} under the same 15-second, 10Hz, high-resolution setting.
On the Waymo dataset, we evaluate on 2,000 validation frames selected from more challenging scenarios. Under the same evaluation setup, DriVerse continues to exhibit clear and consistent improvements over Vista.

\noindent{\textbf{Length-Stability Trade-off. }}
As the generation length increases from 8 to 15 and even 30 seconds, we observe a gradual degradation in FID and FVD, which reflects the common trade-off between sequence length and stability in long-horizon generation. Notably, the most significant drop occurs when moving from 8 to 15 seconds. This is because the 8-second videos are generated in a single forward pass, while longer sequences rely on autoregressive extension, which accumulates errors over time. Nevertheless, DriVerse maintains a clear advantage over prior methods even in long-horizon scenarios, demonstrating strong temporal coherence and robustness under extended predictions.

\noindent{\textbf{Geometric Consistency Evaluation.}}
To evaluate the geometric consistency between the generated video and the target trajectory, we introduce a pose-level alignment error as a structural metric. Specifically, we apply monocular ORB-SLAM~\cite{mur2015orb} to the generated video to estimate a sequence of camera poses $\{\hat{P}_t\}$, which lacks an absolute scale due to the monocular setting. To align these poses with ground-truth camera trajectories $\{P_t\}$, we compute the optimal similarity transformation $(s, R, t)$ that minimizes the mean squared error between the transformed estimated poses and ground truth:
\[
\min_{s, R, t} \sum_{t=1}^{T} \left\| P_t - \left( s \cdot R \cdot \hat{P}_t + t \right) \right\|^2.
\]
This Sim(3) alignment corrects for global scale, rotation, and translation, and the resulting RMSE of aligned trajectories is reported as our \textit{Geometric Alignment Error (GAE)}. A lower GAE indicates better structural agreement between the generated motion and the intended trajectory. Unlike pixel-based perceptual metrics such as FID and FVD, which primarily assess visual fidelity, GAE provides a complementary geometric perspective by directly measuring the model's ability to maintain 3D trajectory fidelity under motion control.

As shown in the upper part of Table~\ref{tab:waymo_gae}, DriVerse significantly improves trajectory alignment accuracy on the Waymo Open Dataset, reducing the geometric alignment error by approximately 55\% and 44\% at 8-second and 15-second settings respectively, compared to the Vista baseline. As the generation length increases, the error gradually rises, reflecting the accumulation of trajectory drift in long-horizon prediction.

\noindent{\textbf{Qualitative Results.}}
As shown in Figure~\ref{fig:e1}, we directly apply our DriVerse model trained on Waymo to a visualization frame from the original Vista paper. Without any fine-tuning, DriVerse generates high-quality videos twice as long as Vista’s, demonstrating strong few-shot generalization. Additionally, qualitative results on the nuScenes validation set highlight DriVerse’s superior semantic generation capability in complex driving scenes.

\begin{table}[t]
\footnotesize
\centering
\caption{Geometric Alignment Error (GAE) comparison on Waymo Open Dataset. Max Video Length denotes the temporal range of generation. Lower is better. Bold indicates the best performance. }
\vspace{-1.2em}
\setlength{\tabcolsep}{4.2mm}
\begin{tabular}{lcc}
\toprule
Method & Max Video Length & GAE $\downarrow$ (m) \\
\midrule
Vista & 10Hz, 15s & 3.72 \\
\rowcolor{gray!10}
DriVerse  & 10Hz, 8s & \textbf{1.68} \\
\rowcolor{gray!10}
DriVerse  & 10Hz, 15s & 2.10 \\
\rowcolor{gray!10}
DriVerse  & 10Hz, 30s & 2.74 \\
\bottomrule
\end{tabular}
\label{tab:waymo_gae}
\vspace{-1.2em}
\vspace{-6pt}
\end{table}

\begin{table}[t]
\footnotesize
\centering
\caption{Ablation study on the impact of MTP, LMA, and DWG modules based on DriVerse (10Hz, 8s) on the Waymo Open Dataset. Lower GAE indicates better geometric alignment. Bold indicates the best performance. }
\setlength{\tabcolsep}{3.8mm}
\begin{tabular}{c|ccc|c}
\toprule
Setting & MTP & LMA & DWG & GAE $\downarrow$ (m) \\
\midrule
\rowcolor{gray!10}
(a)  & \checkmark & \checkmark & \checkmark & \textbf{1.68} \\
(b)   & -          & \checkmark & \checkmark & 3.25 \\
(c)  & \checkmark & -          & \checkmark & 1.73 \\
(d) & \checkmark & \checkmark & -          & 2.01 \\
\bottomrule
\end{tabular}
\label{tab:ablation_gae}
\end{table}

\begin{table}[t]
\footnotesize
\centering
\caption{Comparison under scenarios with heading angle change exceeding 40° on Waymo Open Dataset, evaluated at 10Hz, 15s. Bold indicates the best. }
\setlength{\tabcolsep}{3.4mm}
\begin{tabular}{l|ccc|c}
\toprule
Method & FID $\downarrow$ & FVD $\downarrow$ & GAE $\downarrow$ (m) \\
\midrule
Vista & 38.43 & 422.4 & 8.48 \\
DriVerse (Full) & \textbf{35.34} & \textbf{335.6} & \textbf{4.65} \\
DriVerse w/o DWG & 36.55 & 395.1 & 7.54 \\
\bottomrule
\end{tabular}
\label{tab:high_heading_ablation}
\end{table}

\subsection{Ablation Study}
\noindent{\textbf{Effectiveness of Multimodal Trajectory Prompting.}}
As shown in Table~\ref{tab:ablation_gae}, we conduct an ablation study on the Waymo Open Dataset to evaluate the contribution of the MTP, LMA, and DWG modules based on our DriVerse model (10Hz, 8s). Notably, in setting (b), we replace our MTP trajectory encoding with a direct 3D-to-2D projection of the trajectory onto the image plane. This substitution leads to a significant performance drop to 3.25 meters, highlighting the effectiveness of our method in enhancing pixel-level control and alignment. In setting (c), removing LMA only causes a minor degradation. This is primarily because the influence of dynamic agents is less prominent in scenes dominated by static elements, making its effect less observable in the global metric. Similarly, removing DWG in setting (d) results in a modest increase in error, which we attribute to the relatively low proportion of scenes involving large heading angle changes in the validation set. We further validate the hypothesis in later analysis.

\noindent{\textbf{Effectiveness of Latent Motion Alignment.}}
Although the impact of the LMA module on quantitative metrics is not as prominent, its effectiveness becomes visually evident when comparing qualitative results. As shown in Figure~\ref{fig:e2}, the top two rows demonstrate the ability of DriVerse to generate diverse and plausible futures (Future A/B). When the LMA module is removed (third row), we observe clear motion anomalies such as vehicles reversing or making abrupt direction changes in intersections, as highlighted by the red dashed circles. This indicates that the lack of LMA significantly weakens the model's ability to understand object-level motion semantics.

\noindent{\textbf{Effectiveness of Dynamic Window Generation.}}
As shown in Table~\ref{tab:high_heading_ablation}, we evaluate our model under challenging scenarios where the heading angle change exceeds $40^\circ$, using a frame rate of 10Hz and a duration of 15 seconds. Compared to Vista, our full DriVerse model achieves significantly better performance across all metrics, reducing GAE from 8.48m to 4.65m, along with notable improvements in FID and FVD. When the DWG module is removed, the performance degrades substantially (e.g., GAE increases to 7.54m), demonstrating the critical role of DWG in handling large rotational motions. These results validate the effectiveness of DWG, particularly its “when things go wrong, early do the next regression” strategy, in complex and high-dynamic scenarios, confirming our earlier hypothesis.

\noindent{\textbf{Feasibility of Directly Using a Base Video Generation Model.}} As shown in the bottom row of Figure~\ref{fig:e2}, we further examine a base image-to-video generation model conditioned solely on textual descriptions, without any explicit control signals. The generated results exhibit severe visual artifacts, including object splitting, vanishing, and blending with the background, as marked by white dashed circles and arrows. This supports our earlier discussion: base video generation models, which are often trained on datasets with limited motion and few subjects, struggle to generalize to complex outdoor scenes—especially those with multiple agents and large-scale dynamic motion, as in autonomous driving scenarios.

\section{Conclusion}
We propose DriVerse, a world simulator for autonomous driving that generates future video sequences conditioned on a single input image and a given trajectory. To enable accurate modeling of both camera motion (static content) and object motion (dynamic content), we introduce two modules: Multimodal Trajectory Prompting (MTP) and Latent Motion Tracking (LMT). These modules empower a base video diffusion model—originally trained on short-range, low-motion scenes—to rapidly adapt to long-range, complex urban street views with diverse dynamics, using only limited finetuning. Furthermore, we present a Dynamic Window Generation (DWG) strategy to address degradation in generation quality under sharp changes in vehicle heading. Experimental results show that DriVerse achieves state-of-the-art performance in both visual quality and trajectory alignment, producing videos with high realism and spatial consistency. Our method offers an effective new paradigm for future scene simulation in autonomous driving.

\noindent\textbf{Limitation.} While DriVerse focuses on simulating future scenes conditioned on a given trajectory via video generation, and proposes a novel geometric consistency metric by reconstructing trajectories from generated videos, it has not yet been integrated into a full-stack autonomous driving training loop or simulator. In future work, we aim to: (1) explore DriVerse as a trajectory evaluator or reward signal in end-to-end driving models, and (2) combine DriVerse with existing implicit urban scene reconstruction frameworks to improve simulation quality in complex environments.

{\small
\bibliographystyle{ieee_fullname}
\bibliography{main}
}
\end{document}